\documentclass[runningheads]{llncs}
\usepackage{graphicx}
\usepackage{amsmath}
\usepackage{booktabs}
\usepackage{graphicx}
\usepackage{times}
\usepackage{epsfig}
\usepackage{dsfont}
\usepackage{cite}
\usepackage{amsmath}
\usepackage{amssymb}
\usepackage{multirow}
\usepackage[normalem]{ulem}
\usepackage{url}
\usepackage{xcolor}
\usepackage{dirtytalk}
\usepackage{hyperref}
\usepackage[normalem]{ulem}
\useunder{\uline}{\ul}{}
\usepackage[subrefformat=parens,labelformat=parens]{subcaption}

\newcommand\samethanks[1][\value{footnote}]{\footnotemark[#1]}
\usepackage{pdfpages}

\begin{document}
\title{Self-supervised Mean Teacher for Semi-supervised Chest X-ray Classification }


\author{
Fengbei Liu \inst{1\thanks{First two authors contributed equally to this work.}} $\quad$
Yu Tian\inst{1,4\samethanks}$\quad$
Filipe R. Cordeiro\inst{2} $\quad$
Vasileios Belagiannis \inst{3}$\quad$ \\
Ian Reid\inst{1}$\quad$
Gustavo Carneiro\inst{1}
}

% \author{Paper ID\# ???}

% \authorrunning{Paper ID\# ???}
\institute{Australian Institute for Machine Learning, University of Adelaide \and
Universidade Federal Rural de Pernambuco, Brazil \and
Universit\"at Ulm, Germany \and
South Australian Health and Medical Research Institute
}
\maketitle              % typeset the header of the contribution
\vspace{-.3in}
\begin{abstract}

The training of deep learning models generally requires a large amount of annotated data for effective convergence and generalisation.
However, obtaining high-quality annotations is a laboursome and expensive process due to the need of expert radiologists for the labelling task.
The study of semi-supervised learning in medical image analysis is then of crucial importance given that it is much less expensive to obtain unlabelled images than to acquire images labelled by expert radiologists. Essentially, semi-supervised methods leverage large sets of unlabelled data to enable better training convergence and generalisation than using only the small set of labelled images. In this paper, we propose Self-supervised Mean Teacher for Semi-supervised (S$^2$MTS$^2$) learning that combines self-supervised mean-teacher pre-training with semi-supervised fine-tuning. The main innovation of S$^2$MTS$^2$ is the self-supervised mean-teacher pre-training based on the joint contrastive learning, which uses an infinite number of pairs of positive query and key features to improve the mean-teacher representation. The model is then fine-tuned using the exponential moving average teacher framework trained with semi-supervised learning. We validate S$^2$MTS$^2$ on the multi-label classification problems from Chest X-ray14 and CheXpert, and the multi-class classification from ISIC2018, where we show that it outperforms the previous SOTA semi-supervised learning methods by a large margin. Our code is available at \url{https://github.com/fengbeiliu/semi-chest}. 

\end{abstract}
\vspace{-.3in}
\section{Introduction}
\vspace{-.1in}
\label{sec:introduction}

Deep learning has shown outstanding results in medical image analysis problems~\cite{lee2017deep,tian2020few,li2018thoracic,tian2019one,liu2020self,jonmohamadi2020automatic}. 
However, this performance usually depends on the availability of labelled datasets, which is expensive to obtain given that the labelling process requires expert doctors.
This limitation motivates the study of semi-supervised learning (SSL) methods that train models with a small set of labelled data and a large set of unlabelled data. 

The current state-of-the-art (SOTA) SSL is based on 
pseudo-labelling methods~\cite{rizve2021defense,lee2013pseudo}, consistency-enforcing approaches~\cite{berthelot2019remixmatch,tarvainen2017mean,laine2016temporal}, self-supervised and semi-supervised learning (S$^4$L)~\cite{chen2020big,zhai2019s4l}, and graph-based label propagation~\cite{aviles2019graphx}.  Pseudo-labelling is an intuitive SSL technique, where confident predictions from the model are transformed into pseudo-labels for the unlabelled data, which are then used to re-train the model~\cite{lee2013pseudo}.
%Pseudo-labelling alone does not produce SOTA results~\cite{oliver2018realistic}, but its effectiveness increases when combined with other methods~\cite{pham2019semi}.  
Consistency-enforcing regularisation is based on training for a consistent output given model~\cite{liu2020semi,tarvainen2017mean} or input data~\cite{berthelot2019remixmatch,laine2016temporal} perturbations. 
S$^4$L methods are based on self-supervised pre-training~\cite{moco,simclr}, followed by supervised fine-tuning using few labelled samples~\cite{chen2020big,zhai2019s4l}.
Graph-based methods rely on label propagation on graphs~\cite{aviles2019graphx}.
Recently, Yang et al.~\cite{yang2020rethinking} suggested that self-supervision pre-training provides better feature representations than consistency-enforcing approaches in SSL.  
However, previous S$^4$L approaches use only the labelled data in the fine-tuning stage, missing useful training information present in the unlabelled data. 
Furthermore, self-supervised pre-training~\cite{simclr,moco} tends to use limited amount of samples to represent each class, but recently, Cai et al.~\cite{cai2020joint} showed that better representation can be obtained with an infinite amount of samples.
Also, recent research~\cite{rizve2021defense} suggests that the student-teacher framework, such as the mean-teacher~\cite{tarvainen2017mean}, works better in multi-label semi-supervised tasks
than other SSL methods.
We speculate that this is because other methods are usually designed to work with softmax activation that only works in multi-class problems, while mean-teacher~\cite{tarvainen2017mean} does not have this constraint and can work in multi-label problems.  

In this paper, we propose a self-supervised mean-teacher for semi-supervised (S$^2$MTS$^2$) learning approach that combines S$^4$L~\cite{chen2020big,zhai2019s4l} with consistency-enforcing learning based on the mean-teacher algorithm~\cite{tarvainen2017mean}.
The main contribution of our method is the self-supervised mean-teacher pre-training with the joint contrastive learning~\cite{cai2020joint}. To the best of our knowledge, this is the first approach, in our field, to train the mean teacher model with self-supervised learning.
%\filipe{We should make it clearer that the self-supervised mean teacher is something new. Otherwise it seems it's just combining self-supervision and semi-supervision, but we propose a new self-supervised method, which we combine later with the semi-supervision.}
%To the best of our knowledge, this is the first time such combination is proposed for solving semi-supervised learning challenges.
%During the self-supervised pre-training, we adapt the joint contrastive learning~\cite{cai2020joint}, which uses an infinite number of pairs of positive query and key features, to efficiently train a mean teacher model with unsupervised learning.
%we rely on an infinite number of pairs of positive query and key features~\cite{cai2020joint}, efficiently sampled from a Gaussian distribution learned from the augmented unlabelled data, to train a momentum based student-teacher model with contrastive learning~\cite{simclr}.
This model is then fine-tuned with semi-supervised learning using the exponential moving average teacher framework~\cite{tarvainen2017mean}.
We evaluate our proposed method on the thorax disease multi-label datasets ChestX-ray 14~\cite{wang2017chestx} and CheXpert~\cite{irvin2019chexpert}, and on the multi-class skin condition dataset ISIC2018~\cite{tschandl2018ham10000, codella2019skin}.
We show that our method outperforms the SOTA on semi-supervised learning~\cite{aviles2019graphx,liu2020semi,gyawali2020semi,unnikrishnan2020semi}. Moreover, we investigate each component of our framework for their contribution to the overall model in the ablation study. 

% We evaluate S$^2$MTS$^2$ on the Chest X-ray14 dataset~\cite{wang2017chestx} and CheXpert dataset \cite{irvin2019chexpert} that contains thorax disease multi-label classification. In the benchmark that uses up to 20\% of the training set as labelled,
% we show that our method outperforms the SOTA on semi-supervised learning~\cite{aviles2019graphx,liu2020semi}. Moreover, we investigate each component of our framework for their contribution to the overall model in the ablation study. 
% In conclusion, Out contributions include: \\
% \begin{itemize}
%     \item 
% \end{itemize}
% We validate S$^2$MTS$^2$ on the thorax disease multi-label classification problem from the dataset Chest X-ray14, where we show that it outperforms the previous state-of-the-art semi-supervised learning methods~\cite{aviles2019graphx,liu2020semi} by a large margin. \gustavo{the text pretty much repeats the abstract, so it needs to be changed.  I need to move to the new sections for the meeting, so if you have time, please go ahead and make the changes.}

\begin{figure}[t!]
\begin{center}

% \includepdf[pages=4,width=\textwidth,pagecommand=\subsection{blub}]{MICCAI Chest semi.pdf}
\includegraphics[width=0.96\textwidth,page=4]{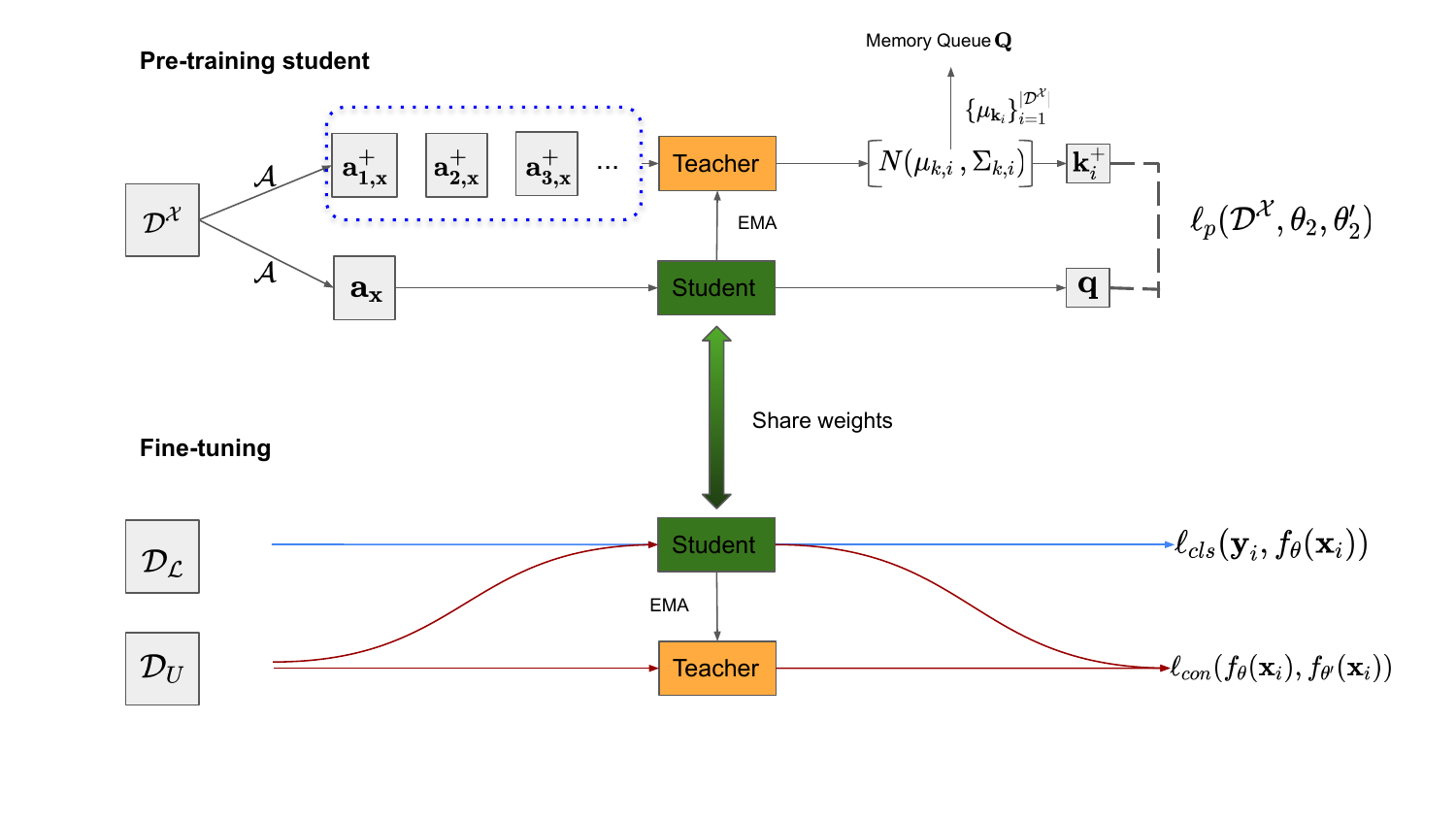}
\end{center}
\vspace{-.5in}
  \caption{Description of the proposed 
  self-supervised mean-teacher for semi-supervised (S$^2$MTS$^2$) learning.
  The main contribution of the paper resides in the top part of the figure, with the self-supervised mean-teacher pre-training based on joint contrastive learning, which uses an infinite number of pairs of positive query and key features sampled from the unlabelled images to minimise $\ell_p(.)$ in~\eqref{eq:loss_pre_train}.
  This model is then fine-tuned with the exponential moving average teacher in a semi-supervised learning framework that uses both labelled and unlabelled sets to minimise $\ell_{cls}(.)$ and $\ell_{con}(.)$ in~\eqref{eq:loss_fine_tune}.
   %pre-training stage to pre-train the student network to generate good representations for student network, and a fine-tuning stage to fine-tune the student network with labelled and unlabelled data. \gustavo{Make sure the letters, loss names follow what is in the paper. Right now, there are some differences that need to be fixed in this figure. Please increase font size.} \yu{would be better if we use figure from real x-ray images}\fengbei{I will revisit this figure to change loss name and add some figures for the augmentation}. \filipe{please make clear that the pretraining is the self-supervised stage. Also, you should say if X in the top figure is the labelled or unlabelled set.\fengbei{Top figure X always unlabel because self-supervised?}}
   } 
%   \vspace{-.1in}
\label{fig:framework}
\end{figure}

\vspace{-.1in}
\section{Related Works}

SSL is a research topic that is gaining attention from the medical image analysis community due to the expensive image annotation process~\cite{cheplygina2019not} and the growing number of large-scale datasets available in the field~\cite{wang2017chestx}. 
%Techniques based on self-labelling, which iteratively update the unlabelled data with pseudo-labels has been adopted by various methods~\cite{bai2019self,singh2011identifying}. 
%The use of generative models to augment labelled and unlabelled data has been explored for SSL in~\cite{diaz2019retinal}. 
%GraphxNet~\cite{aviles2019graphx} relies on a graph to propagate labels from the labelled to the unlabelled images. 
The current SOTA SSL methods are based on consistency-enforcing approaches that leverage the unlabelled data to regularise the model prediction consistency~\cite{tarvainen2017mean,laine2016temporal}. 
 Other related papers~\cite{cui2019semi} extend the mean teacher~\cite{tarvainen2017mean} to encourage consistency between the prediction by the student and teacher models for atrium and brain lesion segmentation. 
The SOTA SSL method on Chest X-ray images~\cite{liu2020semi} exploits the consistency in the relations between labelled and unlabelled data. 
None of these methods explores a self-supervised consistency-enforcing  method to pre-train an SSL model, as we propose.

Self-supervised learning methods~\cite{simclr,moco} are also being widely investigated in SSL because they can provide good representations~\cite{chen2020big,zhai2019s4l}. However, these methods ignore the large amount of unlabelled data to be used during SSL, which may lead to unsatisfactory generalisation process. %(In Table \ref{tab:ablation} 
An important point in self-supervised learning is on how to define the classes to be learned.
In general, each class is composed of a single pair of augmented images from the same image, and many pairs of augmentations from different images~\cite{moco,simclr,cai2020joint,chen2020big}.
The use of a single pair of images to form a class has been criticised by Cai et al.~\cite{cai2020joint}, who propose the joint contrastive learning (JCL), which is an efficient way to form a class with an infinite number of augmented images from the same image to leverage the statistical dependency between different augmentations.

%\gustavo{Talk about the specific SSL methods for the Chest X-ray14~\cite{aviles2019graphx,liu2020semi}. Mention their disadvantages with respect to our method.

%Talk about self-supervised methods~\cite{simclr,moco}, and explain the advantage of~\cite{cai2020joint}.Talk about S$^4$L approaches~\cite{chen2020big,dai2015semi,zhai2019s4l}.}

\section{Method}

%Figure~\ref{fig:framework} demonstrate our proposed contratsive feature augmentation learning (CSAL) pretraining for unlabelled chestXray data. 

In this section, we introduce our two-stage learning framework in detail (see Fig.~\ref{fig:framework}). 
We assume that we have a small labelled dataset, denoted by $\mathcal{D}_L = \{ (\mathbf{x}_i,\mathbf{y}_i)\}_{i=1}^{|\mathcal{D}_L|}$, where the image is represented by $\mathbf{x} \in \mathcal{X} \subset \mathbb{R}^{H \times W \times C}$, and class 
$\mathbf{y} \in \{0,1\}^{|\mathcal{Y}|}$, where $\mathcal{Y}$ represents the label set.
%the \fengbei{attribute} \yu{attribute set is confusing though, especially for those that are not familiar with the multi-label field } set. 
We consider a multi-label problem and thus $\sum_{c=1}^{|\mathcal{Y}|}\mathbf{y}_i(c) \in [0,|\mathcal{Y}|]$. The unlabelled dataset is defined by $\mathcal{D}_U = \{ \mathbf{x}_i\}_{i=1}^{|\mathcal{D}_U|}$ with $|\mathcal{D}_L| << |\mathcal{D}_U|$.

Our model consists of a  student and a teacher model~\cite{tarvainen2017mean}, denoted by parameters $\theta,\theta' \in \mathbf{\Theta}$, respectively, which parameterize the classifier $f_{\theta}:\mathcal{X} \rightarrow [0,1]^{|\mathcal{Y}|}$.
This classifier can be decomposed as $f_{\theta} = h_{\theta_1}\circ g_{\theta_2}$, with $g_{\theta_2}:\mathcal{X} \rightarrow \mathcal{Z}$ and $h_{\theta_1}:\mathcal{Z} \rightarrow [0,1]^{|\mathcal{Y}|}$.
The first stage (top of Fig.~\ref{fig:framework}) of the training consists of a self-supervised learning that uses the images from $\mathcal{D}_L$ and $\mathcal{D}_U$, denoted by $\mathcal{D}^{\mathcal{X}} = \{ \mathbf{x}_i | \mathbf{x}_i \in \mathcal{D}^{\mathcal{X}}_L \bigcup \mathcal{D}_U\}_{i=1}^{|\mathcal{D}_{\mathcal{X}}|}$, with $\mathcal{D}^{\mathcal{X}}_L$ representing the images from the set $\mathcal{D}_L$, where our method minimises the joint contrastive learning loss~\cite{cai2020joint}, defined in~\eqref{eq:loss_pre_train}.  This means that during this first stage, we only learn the parameters for $g_{\theta_2}$.
The second stage (bottom of Fig.~\ref{fig:framework}) fine-tunes this pre-trained student-teacher model using the semi supervised consistency loss defined in~\eqref{eq:loss_fine_tune}. Below we provide details on the losses and training.

%The images are sampled from the data distribution $\mathcal{X}$, and we represent the labelled subset as $\mathcal{D}_L = \{ \mathbf{x}_i\}_{i=1}^{|\mathcal{D}_L|}$, where the image is represented by $\mathbf{x} \in \mathcal{X} \subset \mathbb{R}^{H \times W \times C}$, and class $y \in \mathcal{Y}$, where $\mathcal{Y}$ represents the one hot multi-label ground truth. Below, we explain how to utilise self-supervised pretraining to enhance the performance of semi-supervised tasks.

%\subsection{Contratsive feature augmentation learning}
\vspace{-.1in}
\subsection{Joint Contrastive Learning to Self-supervise the Mean-teacher Pre-training}

% We derive a particular form of contrastive loss named Joint Contrastive Learning (JCL). JCL implicitly involves the simultaneous learning of an infinite number of query-key pairs, which poses tighter constraints when searching for invariant features

The self-supervised pre-training of the mean-teacher using joint contrastive learning (JCL)~\cite{cai2020joint}, presented in this section, is the main technical contribution of this paper. 
%\filipe{I agree. I think we should emphasise that in the abstract and introduction. Otherwise it seems we are just combining self-supervision with semi-supervision.}
The teacher and student process an input image to return the keys $\mathbf{k}\in\mathcal{Z}$ and the queries $\mathbf{q}\in\mathcal{Z}$ with $\mathbf{k}=g_{\theta'_2}(\mathbf{x})$ and $\mathbf{q}=g_{\theta_2}(\mathbf{x})$. We also assume that we have a set of augmentation functions, i.e.,~random crop and resize, rotation and Gaussian blur, denoted by $\mathcal{A}=\{ a_l:\mathcal{X} \rightarrow \mathcal{X} \}_{l=1}^{|\mathcal{A}|}$.    
%The self-supervised pretraining aims to learn good representations for downstream tasks in a task-agnostic way via unsupervised learning (i.e., no labels)~\cite{moco,simlar}.
%Recent work utilise contrastive learning to attract the query-key feature positive pairs and repell that of the negative pairs to form an auxiliary uniform 'multi-class' distributions~\cite{moco,simlar,wang2020understanding,cai2020joint,chen2020big}. Each image sample is considered as a single class and the augmentations(views) of that images belongs to the distribution of this class. The equation to achieve this is through a
Then JCL minimises the following loss~\cite{cai2020joint}:
\begin{equation}
    {
    \ell_{p}(\mathcal{D}^{\mathcal{X}},\theta_2,\theta'_2)=
    -\frac{1}{|\mathcal{D}^{\mathcal{X}}|}
    \frac{1}{M}
    \sum_{i=1}^{|\mathcal{D}^{\mathcal{X}}|}
    \sum_{m=1}^{M}
    \left [
    \log\frac{\exp\left[ \frac{1}{\tau} \mathbf{q}_i^{\top} \mathbf{k}_{i,m}^{+} \right]}{\exp\left[ \frac{1}{\tau} \mathbf{q}_i^{\top} \mathbf{k}_{i,m}^{+} \right] + \sum_{j=1}^K \exp\left[ \frac{1}{\tau}  \mathbf{q}_i^{\top} \mathbf{k}_{i,j}^{-} \right]}
    \right ],}
    \label{eq:loss_pre_train}
\end{equation}
where $\tau$ is the temperature hyper-parameter, the query $\mathbf{q}_i = g_{\theta_2}(a(\mathbf{x}_i))$, with $a \in \mathcal{A}$.
the positive key $\mathbf{k}_{i,m}^{+} \sim p(\mathbf{k}_i^{+})$, with $p(\mathbf{k}_i^{+}) = \mathcal{N}(\mu_{\mathbf{k}_i},\Sigma_{\mathbf{k}_i})$ and $\mathbf{k}_i = g_{\theta'_2}(a(\mathbf{x}_i))$ (i.e., a sample from the data augmentation distribution for $\mathbf{x}$), and the negative keys $\mathbf{k}_{i,j}^{-} \in \{ \mu_{\mathbf{k}_j} \}_{i,j \in \{1,..,|\mathcal{D}^{\mathcal{X}}|\}, i \ne j }$ represents a negative key for query $\mathbf{q}_i$. 
In~\eqref{eq:loss_pre_train}, $M$ denotes the number of positive keys, and Cai et al.~\cite{cai2020joint} describe a loss that minimises a bound to~\eqref{eq:loss_pre_train} for $M \to \infty$ -- below, the minimisation of $\ell_p(.)$ in~\eqref{eq:loss_pre_train} is realised by the minimisation of this bound.
As defined above, the generative model $p(\mathbf{k}_i^{+})$ is denoted by the Gaussian 
$\mathcal{N}(\mu_{\mathbf{k}_i},\Sigma_{\mathbf{k}_i})$, where the mean $\mu_{\mathbf{k}_i}$ and covariance $\Sigma_{\mathbf{k}_i}$ are estimated from a set of keys $\{ \mathbf{k}^{+}_{i,l} = g_{\theta'_2}(a_l(\mathbf{x}_i))  \}_{a_l \in \mathcal{A}}$ formed by different views of $\mathbf{x}_i$.
The set of negative keys $\{ \mu_{\mathbf{k}_j} \}_{i,j \in \{1,..,|\mathcal{D}^{\mathcal{X}}|\}, i \ne j }$ is stored in a memory queue~\cite{moco} that is updated in a first-in-first-out way, where the mean of the keys in $\{ \mu_{\mathbf{k}_i} \}_{i=1}^{|\mathcal{D}^{\mathcal{X}}|} $ are inserted to the memory queue to replace the oldest key means from previous training iterations. The memory queue has been designed to increase the number of negative samples without sacrificing computation efficiency. 

The training of the student-teacher model~\cite{moco,zhou2020comparing,tarvainen2017mean} is achieved by updating the student parameter using the loss in~\eqref{eq:loss_pre_train}, as in
$\theta_2(t) = \theta_2(t-1) - \nabla_{\theta_2}\ell_{p}(\mathcal{D}^{\mathcal{X}},\theta_2,\theta'_2)$, where $t$ is the training iteration.  The teacher model parameter is updated with exponential moving average (EMA) with $\theta'_2(t) = \alpha \theta'_2(t-1) + (1-\alpha)\theta_2(t) $, with $\alpha\in[0,1]$.
For this pre-training stage, we notice that training for more epochs always improve the model regularisation given that it is difficult to overfit the training set with the loss in~\eqref{eq:loss_pre_train}. 
Hence, we select the last epoch student model $g_{\theta_2}(.)$ to initialise the fine-tuning stage, defined below in Sec.~\ref{sec:fine_tune}.

\subsection{Fine-tuning the Mean Teacher}
\label{sec:fine_tune}

To fine tune the mean teacher, we follow the approach in~\cite{moco,tarvainen2017mean} using the following loss to train the student model:
\begin{equation}
    %\ell_{t}(\mathcal{D}_L,\mathcal{D}_U,\theta,\theta') = -\frac{1}{|\mathcal{D}_L|}\sum_{(\mathbf{x}_i,\mathbf{y}_i) \in \mathcal{D}_L} \mathbf{y}_i^{\top}\log(f_{\theta}(\mathbf{x}_i)) + \frac{1}{|\mathcal{D}|}\sum_{\mathbf{x}_i \in \mathcal{D}} \| f_{\theta}(\mathbf{x}_i) - f_{\theta'}(\mathbf{x}_i) \|^2,
    \ell_{t}(\mathcal{D}_L,\mathcal{D}_U,\theta,\theta') = \frac{1}{|\mathcal{D}_L|}\sum_{(\mathbf{x}_i,\mathbf{y}_i) \in \mathcal{D}_L} \ell_{cls}(\mathbf{y}_i,f_{\theta}(\mathbf{x}_i))  + \frac{1}{|\mathcal{D}|}\sum_{\mathbf{x}_i \in \mathcal{D}} \ell_{con}(f_{\theta}(\mathbf{x}_i),f_{\theta'}(\mathbf{x}_i) ),
    \label{eq:loss_fine_tune}
\end{equation}
%\fengbei{Shall we separate two loss term and give them name. So I can use this name in the figure for each labelled/unlabelled part?}
where $\ell_{cls}(\mathbf{y}_i,f_{\theta}(\mathbf{x}_i))=-\mathbf{y}_i^{\top}\log(f_{\theta}(\mathbf{x}_i))$,
$\ell_{con}(f_{\theta}(\mathbf{x}_i),f_{\theta'}(\mathbf{x}_i) ) = \| f_{\theta}(\mathbf{x}_i) - f_{\theta'}(\mathbf{x}_i) \|^2$, and 
$\mathcal{D}=\mathcal{D}_U \bigcup \mathcal{D}^{\mathcal{X}}_L$.
The training of the student-teacher model~\cite{moco,tarvainen2017mean,zhou2020comparing} is achieved by updating the student parameter using the loss in~\eqref{eq:loss_fine_tune}, as in
$\theta(t) = \theta(t-1) - \nabla_{\theta}\ell_{t}(\mathcal{D}_L,\mathcal{D}_U,\theta,\theta')$, where $t$ is the training iteration.  The teacher model parameter is updated with exponential moving average (EMA) with $\theta'(t) = \alpha \theta'(t-1) + (1-\alpha)\theta(t) $, with $\alpha\in[0,1]$.
After finishing the fine-tuning stage, we select the teacher model $f_{\theta'}(.)$ to estimate the multi-label classification for test images.

\vspace{-.1in}
\section{Experiment}
\vspace{-.1in}
\subsection{Dataset Setup}
We use Chest X-ray14~\cite{wang2017chestx}, CheXpert~\cite{irvin2019chexpert} and ISIC2018~\cite{tschandl2018ham10000, codella2019skin} datasets to evaluate our method. \\
\textbf{Chest X-ray14} contains 112,120 chest x-ray images from 30,805 different patients. There are 14 different labels (each label represents a disease) in the dataset, where each patient can have multiple diseases at the same time, forming a multi-label classification problem. To compare with previous papers~\cite{aviles2019graphx,liu2020semi}, we adopt the official train/test data split. For the self-supervised pre-training of the mean teacher, we used all the unlabelled images (86k samples) from the training set. For the semi-supervised fine-tuning of the mean teacher, we follow the papers~\cite{aviles2019graphx,liu2020semi} and experiment with training sets containing different proportions of labelled data (2\%,5\%,10\%,15\%,20\%). We report the classification result on the official test set (26,000 samples) using area under the ROC curve (AUC).\\
\textbf{CheXpert} contains around 220,000 images with 14 different diseases, and similarly to Chest X-ray14, each patient can have multiple diseases at the same time. For pre-processing, we remove all lateral view images and treat uncertain label as negative labels. We follow the semi-supervised setup from~\cite{gyawali2020semi}, and experiment with 100/200/300/ 400/500 labelled samples per class. %\gustavo{how can you have 500k labelled samples with 220k images?}. 
We report results on the official test set using AUC.\\
\textbf{ISIC2018} is a multi-class skin condition dataset that contains 10,015 images with seven different labels. Each image is associated with one of the seven labels, forming a multi-class classification problem. We follow~\cite{liu2020semi} train/test split for fair comparison, where the training contains 20\% of the samples labelled, and the remaining 80\% unlabelled. We report the AUC, Sensitivity, and F1 score results to compare with baselines.

% We use Chest X-ray14 ~\cite{wang2017chestx}, CheXpert \cite{irvin2019chexpert} and ISIC2018 ~\cite{tschandl2018ham10000, codella2019skin} to evaluate our method. The dataset contains 112,120 chest x-ray images from 30,805 different patients. There are 14 different labels (each label represents a disease) in the dataset, where each patient can have multiple diseases at the same time, forming a \textbf{multi-label} classification problem. To compare with previous papers~\cite{aviles2019graphx,liu2020semi}, we adopt the official train/test data split. For the self-supervised pre-training of the mean teacher, we used all the unlabelled images (86k samples) from the training set. For the semi-supervised fine-tuning of the mean teacher, we follow the papers~\cite{aviles2019graphx,liu2020semi} and experiment with training sets containing different proportions of labelled data (2\%,5\%,10\%,15\%,20\%). We report the classification result on the official test set (26k samples) using area under the ROC curve (AUC).

\subsection{Implementation Details}

For all datasets, we use DenseNet121~\cite{huang2017densely} as our backbone model. For self-supervised pre-training, we follow~\cite{chen2020big} and replace the two-layer multi-layer perceptron (MLP) projection head by a three-layer MLP.
For dataset pre-processing, we resized Chest X-ray14 images to 512 $\times$ 512 for faster processing and CheXpert and ISIC2018 to 224 $\times$ 224 for fair comparison with baselines. 
We use the data augmentation proposed in~\cite{simclr}, consisting of random resize and crop, random rotation, random horizonntal flipping, except for random grayscale because X-ray images are originally in grayscale.
The batch size is 128 for Chest X-ray14 and 256 for CheXpert and ISIC2018, and learning rate is 0.05. For the fine-tuning stage, we use batch size 32 with 16 labelled and 16 unlabelled. The fine-tuning takes 30 epochs with learning rate decayed by 0.1 at 15 and 25 epochs for all datasets. We use the SGD optimiser with 0.9 momentum for the pre-training stage, and Adam optimiser in fine-tuning stage. The code is written in Pytorch ~\cite{paszke2019pytorch}. 
We use 4 Nvidia Volta-100 for the self-supervised stage and 1 Nvidia RTX 2080ti for fine-tuning.

% For all dataset, We use the DenseNet-121 \cite{huang2017densely}
% as our backbone model. We use Adam optimiser and learning rate of 0.03. 
% Furthermore,  we follow Chen et al.'s  discovery~\cite{chen2020big} and replace the projection head from a two-layer multi-layer perceptron (MLP)~\cite{simclr} to a three-layer MLP. 
% similarly to~\cite{hermoza2020region,wang2017chestx}. 
% We resized the original images from 1024 $\times$ 1024 to 512 $\times$ 512 for faster processing. 
% For the self-supervised pre-training of the mean teacher, we use the data augmentation proposed in~\cite{simclr}, consisting of random resize and crop, random rotation, and random horizontal flipping, except for the random grayscale because the X-ray images are originally in grayscale. 
% We use Adam optimiser and learning rate of 0.03. 
% Furthermore,  we follow Chen et al.'s  discovery~\cite{chen2020big} and replace the projection head from a two-layer multi-layer perceptron (MLP)~\cite{simclr} 
% %\gustavo{QUESTION: Is this two-layer projection from simclr?} \yu{this is from SIMCLR} 
% to a three-layer MLP. 
% The pre-training stage takes 100 epochs,  

% the batch size is set to 16, and learning rate is 0.05. The fine-tuning takes 30 epochs with learning rate decayed by 0.1 at 15, 25 epochs. The whole code is written in Pytorch. We use 4 Nvidia Volta-100 for the self-supervised stage and 1 Nvidia RTX 2080ti for fine-tuning.
\vspace{-.2in}
\subsection{Experimental Results}
% \usepackage[normalem]{ulem}

%In this section, we show the semi-supervised experimental results on ChestX-ray14. 
We evaluate our approach on the official test set of ChestX-ray14
using different percentage of labelled training data (i.e., 2$\%$, 5$\%$, 10$\%$, 15$\%$, 20$\%$), as shown in Table~\ref{tab:main}. The set of labelled data used for each percentage above follows the same strategy of previous works~\cite{aviles2019graphx,liu2020semi}. 
Our S$^4$L achieves the SOTA AUC results on all different percentages of labels. 
Our model surpasses the previous SOTA SRC-MT~\cite{liu2020semi} by a large margin of 8.7\% and 6.8\% AUC  
for the 2\% and 5\% labelled set cases, respectively, where we use a backbone architecture of lower complexity (Densenet121 instead of the DenseNet169 of~\cite{liu2020semi}). Using the same Densenet121 backbone, GraphXnet~\cite{aviles2019graphx} fails to  classify precisely for the 2\% and 5\% labelled set cases. 
Our method surpasses GraphXnet by more than 20\% AUC in both cases. 
Furthermore, we achieve the SOTA results of the field for the 10\%, 15\% and 20\% labelled set cases, outperforming all previous semi-supervised methods~\cite{liu2020semi,aviles2019graphx}. It is worth noting that our model trained with 5\% of the labelled set achieves better results than SRC-MT with 15\% of labelled.% using a smaller Densenet 121 backbone.  
We also compare with a recently proposed self-supervised pre-training methods, MoCo V2~\cite{chen2020improved}, adapted to our semi-supervised task,  followed by the fine-tuning stage using different percentages of labelled data.
Our method outperforms MoCo V2 by almost 10\% AUC when using 2\% of labelled set, and almost 3\% AUC for 10\% of labelled set. 
Our result for 20\% labelled set achieves comparable 81.06\% AUC performance as the supervised learning approaches -- 81.20\% from MoCo V2 (Densenet 121) and 81.75\% from SRC-MT (Densenet 169) using 100\% of the labelled samples. Such result indicates the effectiveness of our proposed S$^2$MTS$^2$ in SSL benchmark problems. 

We also show the class-level performance using 20\% of the labelled data and compare with other SOTA methods in Tab.~\ref{tab:semi-results}. We compare with the previous baselines, namely original mean teacher (MT) with Densenet169, SRC-MT with Densenet169, MoCo V2, and GraphXNet with Densenet121. We also train a baseline Densenet121 model with 20\% labelled data using Imagenet pre-trained model. Our method achieves the best results on nine classes, surpassing the original MT~\cite{tarvainen2017mean} and its extension SRC-MT~\cite{liu2020semi} by a large margin, demonstrating the effectiveness of our self-supervised learning.

 Furthermore, we compare our approach on the fully-supervised Chest X-ray14 benchmark in Tab.~\ref{tab:supervise-result}. To the best of our knowledge, Hermoza et al.~\cite{hermoza2020region} has the SOTA supervised classification method containing a complex structure (relying on the weakly-supervised localisation of lesions) with a mean AUC of 82.1\% (over the 14 classes), while ours reports a mean AUC of 82.5\%. Hence, our model, using the whole labelled set, achieves the SOTA performance on 8 classes and an average that surpasses the previous supervised methods by a minimum of 0.4\% and a maximum of 8\% AUC. 
% This indicates that our method not only performs well on semi-supervised tasks but also achieves SOTA results on supervised learning.  

%n Fig.~\ref{fig:example_pred}, we show some examples of predictions by our model (trained with 20\% labelled data) on the test set of Chest X-ray14~\cite{wang2017chestx}. 
% %On xx classes, our work achieve the best performance. 

The results on CheXpert and ISIC2018 datasets 
are shown in Tables~\ref{tab:chexpert} and~\ref{tab:isic}, respectively.
%to show the functionality of our model across different datasets. We used the same hyperparameters listed in our paper except for general parameters (batch size, learning etc...) according to image size. 
%On ISIC2018, we follow the SOTA method SRC-MT \cite{liu2020semi} setup with 20\% labeled and 80\% unlabeled split. 
In particular, for CheXpert in Table~\ref{tab:chexpert}, we compare our method with LatentMixing~\cite{gyawali2020semi} and our result is better in all cases. For ISIC2018 on Table~\ref{tab:isic}, using the test set from SRC-MT~\cite{liu2020semi}, our method outperforms all baselines (Supervised, MT, and SRC-MT) for all measures.

\begin{table}[t!]
\centering
\scalebox{0.9}{
\begin{tabular}{c|c|c|c|c|c|c}
\hline
Label Percentage  & 2\%           & 5\%            & 10\%           & 15\%           & 20\%  & 100\% \\ \hline
Graph XNet* \cite{aviles2019graphx}        & 53.00            & 58.00             & 63.00            & 68.00             & 78.00    & N/A   \\
SRC-MT*   \cite{liu2020semi}         & 66.95         & 72.29          & 75.28          & 77.76          & 79.23 & 81.75 \\
NoTeacher \cite{unnikrishnan2020semi}  & 72.60 & 77.04 & 77.61 & N/A &79.49 &N/A\\
MOCO V2~\cite{chen2020improved}     &    65.97      &    73.84     &    77.07      &    79.37       & 80.17 & 81.20  \\ \hline
% Ours (JCL)  \yu{remove this as I put it in ablation?}      & 73.6          & 77.46          & {\ul 79.18}    & {\ul 79.83}    & 80.62 & \#    \\
% Ours (MT vanilla) \yu{remove this as I put it in ablation?} & {\ul 74.8}    & {\ul 77.66}    & 79.08          & 79.7           & \#    &       \\
Ours     & \textbf{74.69} & \textbf{78.96} & \textbf{79.90} & \textbf{80.31} & \textbf{81.06}    &   \textbf{82.51} \\\bottomrule
\end{tabular}
}
\caption{Mean AUC result over the 14 disease classes of Chest X-Ray14 for different label set training percentages. * indicates the methods that use Densenet169 as backbone architecture. \vspace{-.1in}%\fengbei{I have updated NoTeacher result} \
}
\label{tab:main}
\end{table}

% \begin{figure}[t!]
% \begin{center}
% \includegraphics[width= \textwidth]{MICCAI_sample.png}
% \end{center}
%   \caption{Examples of classification by our model (trained with 20\% labelled data) on Chest X-ray14~\cite{wang2017chestx} images. Red indicates the ground truth label.\fengbei{Can we remove these figures, we have to add multiple new tables and there are no space left}}
%   \label{fig:example_pred}
% \end{figure}

% \usepackage{booktabs}
% \usepackage{graphicx}

\begin{table}[t!]
\centering
\scalebox{0.88}{%
\begin{tabular}{@{}c|c|c|c|c|c|c@{}}
\toprule \hline 
Method          & Densenet-121    & GraphXNet ~\cite{aviles2019graphx} & MOCO V2 ~\cite{chen2020improved} & MT ~\cite{liu2020semi} * & SRC-MT ~\cite{liu2020semi} * & Ours \\ \hline  \hline 
Atelectasis     &     75.75                              & 71.89            &     77.21           & 75.12       & 75.38           &       \textbf{78.57}      \\
Cardiomegaly    &     80.71                               & 87.99            &      85.84          & 87.37       & 87.7            &       \textbf{88.08}      \\
Effusion        &     79.87                              & 79.2             &      81.62         & 80.81       & 81.58           &         \textbf{82.87}    \\
Infiltration    &       69.16                           & \textbf{72.05}            &       70.91        & 70.67       & 70.4            &          70.68   \\
Mass            &      78.40                             & 80.9             &       81.71         & 77.72       & 78.03           &      \textbf{82.57}       \\
Nodule          &     74.49                                & 71.13            &      \textbf{76.72}         & 73.27       & 73.64           &        76.60     \\
Pneumonia       &     69.55                                  & \textbf{76.64}            &     71.08           & 69.17       & 69.27           &      72.25       \\
Pneumothorax    &       84.70                            & 83.7             &       85.92         & 85.63       & 86.12           &        \textbf{86.55}     \\
Consolidation   &      71.85                             & 73.36            &       74.47         & 72.51       & 73.11           &        \textbf{75.47}     \\
Edema           &      81.61                              & 80.2             &      83.57          & 82.72       & 82.94           &        \textbf{84.83}     \\
Emphysema       &    89.75                               & 84.07            &      91.10          & 88.16       & 88.98           &         \textbf{91.88}   \\
Fibrosis        &      79.30                              & 80.34            &     80.96           & 78.24       & 79.22           &        \textbf{ 81.73 }   \\
Pleural Thicken &       73.46                             & 75.7             &      75.65          & 74.43       & 75.63           &       \textbf{  76.86 }   \\
Hernia          &      86.05                             & 87.22            &       85.62         & \textbf{87.74 }      & 87.27           &        85.98     \\ \hline 
Mean            &     78.19                               & 78.88            &       80.17         & 78.83       & 79.23           &        \textbf{81.06}     \\ \hline \bottomrule
\end{tabular}%
}
\caption{Class-level AUC comparison between our S$^2$MTS$^2$ and other semi-supervised SOTA approaches trained with \textbf{20\% of labelled data} on Chest X-Ray14. * denotes the methods that use Densenet-169 as backbone.} \vspace{-.2in}
\label{tab:semi-results}
\end{table}

% % Please add the following required packages to your document preamble:
% \usepackage{booktabs}
% \usepackage{graphicx}

\begin{table}[t!]
\centering
\scalebox{0.8}{%
\begin{tabular}{@{}c|c|c|c|c|c|c|c@{}}
\toprule \hline
Method          & Wang et al.~\cite{wang2017chestx} & Li et al.~\cite{li2018thoracic} & CheXNet~\cite{rajpurkar2017chexnet} & CRAL~\cite{guan2020multi} & Ma et al.~\cite{ma2019multi} & Hermoza et al.~\cite{hermoza2020region} & Ours \\ \hline \hline
Atelectasis     & 70          & 72.9      & 75.5    & 78.1 & 77.7      & 77.5           &   \textbf{78.7}   \\
Cardiomegaly    & 81          & 84.6      & 86.7    & 88.3 & \textbf{89.4}     & 88.1           &   87.4   \\
Effusion        & 75.9        & 78.1      & 81.5    & 83.1 & 82.9      & 83.1           &  \textbf{ 83.8 } \\
Infiltration    & 66.1        & 67.3      & 69.4    & 69.7 & 69.6      & 69.5           &   \textbf{70.9}   \\
Mass            & 69.3        & 74.3      & 80.2    & 83   & \textbf{83.8}      & 82.6           &   83.3  \\
Nodule          & 66.9        & 75.8      & 73.5    & 76.4 & 77.1      & 78.9           &   \textbf{79.9}  \\
Pneumonia       & 65.8        & 63.3      & 69.8    & 72.5 & 72.2      & \textbf{74.1}           &   73.9   \\
Pneumothorax    & 79.9        & 79.3      & 82.8    & 86.6 & 86.2      & \textbf{87.9}          &   87.1   \\
Consolidation   & 70.3        & 72        & 72.2    & 75.8 & 75        & 74.7           &   \textbf{75.9}  \\
Edema           & 80.5        & 71        & 83.5    & 85.3 & 84.6      & \textbf{84.6}           &   84.5   \\
Emphysema       & 83.3        & 75.1      & 85.6    & 91.1 & 90.8      & 93.6           &   \textbf{93.7}   \\
Fibrosis        & 78.6        & 76.1      & 80.3    & 82.6 & 82.7      & 83.3           &   \textbf{83.4}   \\
Pleural Thicken & 68.4        & 73        & 74.9    & 78   & 77.9      & 79.3           &   \textbf{79.3}   \\
Hernia          & 87.2        & 66.8      & 89.4    & 91.8 & \textbf{93.4}      & 91.7           &   93.3   \\ \hline
Mean            & 74.5        & 73.9      & 78.9    & 81.6 & 81.7      & 82.1           &   \textbf{82.5}   \\ \hline \bottomrule
\end{tabular}%
}
\caption{Class-level AUC comparison between our S$^2$MTS$^2$ and other supervised SOTA approaches trained with \textbf{100\% of labelled data} on Chest X-Ray14. } \vspace{-.2in}
\label{tab:supervise-result}
\end{table}

\vspace{-.2in}
\begin{table}[t!]
    \begin{minipage}{.5\textwidth}
      \centering
      \scalebox{0.88}{
      \begin{tabular}{c|c|c|c|c|c}
      \toprule Labelled & 100 & 200 & 300 & 400 & 500  \\ \hline
      \midrule LatentMixing \cite{gyawali2020semi} & 65.12 & 66.41 & 67.39 & 67.96& 68.47 \\ \hline
      \bottomrule Ours &  \textbf{66.15} & \textbf{67.85} & \textbf{70.83} & \textbf{71.37} & \textbf{71.58} \\  \bottomrule
      \end{tabular}}
      \caption{Mean AUC result (over the 14 disease classes) on CheXpert for different number of training samples per class.}
      \label{tab:chexpert}
    \end{minipage}
    \begin{minipage}{.5\textwidth}
      \centering
      \scalebox{0.88}{
      \begin{tabular}{c|c|c|c}
      \toprule Method     & AUC   & Sensitivity & F1    \\ \hline
      \midrule Supervised & 90.15 & 65.50       & 52.03 \\ \hline
                MT         & 92.96 & 69.75       & 59.10 \\ \hline
                SRC-MT \cite{liu2020semi}     & 93.58 & 71.47       & 60.68 \\ \hline
      \bottomrule Ours       & \textbf{94.71} & \textbf{72.14}       & \textbf{62.67} \\  \bottomrule
      \end{tabular}}
      \caption{AUC, Sensitivity and F1 result on ISIC2018 using 20\% of labelled training samples.}
      \vspace{-.2in}
      \label{tab:isic}
    \end{minipage}
  \end{table}

% \yu{For 100\% of labels, let's add a table (class-level?) to compared with some other  state-of-the-art methods (fully supervised)... }
\subsection{Ablation Study}

We study the impact of different components of our proposed S$^2$MTS$^2$ in Tab.~\ref{tab:ablation} using Chest X-Ray14. Using the proposed self-supervised learning with just the student model, our model achieves at least 71.95\% mean AUC on various percentages of labelled training data. Adding the JCL component improves the baseline by around 1\% mean AUC on each training percentage. Adding the mean teacher boosts the result by 1.5\% to 2\% mean AUC on each training percentage.
The combination of all our proposed three components achieves SOTA performance on semi-supervised task.

\begin{table}[t!]
\centering
\scalebox{0.88}{
\begin{tabular}{ccc|ccccc}
\toprule\hline
 Self-supervised & JCL & MT    &  AUC (2\%) &  AUC (5\%) &  AUC (10\%) &  AUC (15\%) &  AUC (20\%) \\ \hline \hline
  \checkmark         &      &               &    71.95   &  76.82 &  78.54 &  79.28 &  80.14                \\
      \checkmark         & \checkmark      &               &   72.60    &  77.46 & 79.18  & 79.83 &  80.62             \\

        \checkmark         &      &     \checkmark          &   73.80    &  77.66 & 79.08  & 79.70 &  80.57            \\   \hline
  \checkmark    & \checkmark  & \checkmark       &74.69 &  78.96 &  79.90 &  80.31 & 81.06   \\ \hline\bottomrule
\end{tabular}%
}
\caption{Ablation studies of our method with different components on Chest X-Ray14. "Self-supervised" indicates the traditional self-supervised learning with contrastive loss~\cite{moco}. "JCL" replaces contrastive loss with ~\eqref{eq:loss_pre_train}, "MT" stands for fine-tuned with student-teacher learning instead only fine-tuned on only labelled samples.}\vspace{-.275in}

\label{tab:ablation}
\end{table}

% \begin{table}[t!]
% \centering
% \scalebox{0.88}{
% \begin{tabular}{c|c|c|c|c|c|l}
% \hline
% Label Percentage (k)  & 100           & 200            & 300           & 400           & 500  & \\ \hline
% LatentMixing   \cite{gyawali2020semi}         & 65.12 & 66.41 & 67.39 & 67.96& 68.47
%  \hline
% Ours     &  \textbf{66.15} & \textbf{67.85} & \textbf{70.83} & \textbf{71.37} & \textbf{71.58}
% \end{tabular}
% }
% \caption{Mean AUC result (over the 14 disease classes) for different label set training percentages. * indicates the methods that use Densenet169 as backbone architecture. \fengbei{I have updated NoTeacher result}}
% \label{tab:main}
% \end{table}
\vspace{-.1in}
\section{Conclusion}
\vspace{-.1in}

In this paper, we presented a novel semi-supervised framework, the Self-supervised Mean Teacher for Semi-supervised (S$^2$MTS$^2$) learning.
The main contribution of S$^2$MTS$^2$ is the  self-supervised mean teacher pre-trained based on joint contrastive learning~\cite{cai2020joint}, using an infinite number of pairs of positive query and key features.
This model is then fine-tuned with the exponential moving average teacher framework.
S$^2$MTS$^2$ is validated on the thorax disease multi-label classification problem from the datasets
Chest X-ray14~\cite{wang2017chestx} and CheXpert~\cite{irvin2019chexpert}, and the multi-class classification from the skin condition dataset ISIC2018~\cite{tschandl2018ham10000, codella2019skin}. 
The experiments show that our method outperforms the previous SOTA semi-supervised learning methods by a large margin in all benchmarks containing a varying percentage of labelled data. 
We also show that the method holds the SOTA results on Chest X-ray14~\cite{wang2017chestx} even for the fully-supervised problem.
The ablation study shows the importance of three main components of the method, namely self-supervised learning, JCL, and the mean-teacher model. 
We will investigate the performance of our method on other semi-supervised medical imaging benchmarks in the future.

\textbf{Acknowledgement} This work was supported by Australian Research Council through grants DP180103232 and FT190100525.

\bibliographystyle{splncs04}
% \bibliographystyle{IEEEbib}
\bibliography{bibli}

\end{document}